%% file: main.tex
\documentclass{article}

\usepackage[final]{neurips_2024}

\usepackage[utf8]{inputenc} 
\usepackage[T1]{fontenc}    
\usepackage{hyperref}       
\usepackage{url}            
\usepackage{booktabs}       
\usepackage{amsfonts}       
\usepackage{nicefrac}       
\usepackage{microtype}      
\usepackage{algorithm}
\usepackage[noend]{algpseudocode}
\usepackage[table, xcdraw]{xcolor}
\usepackage{graphicx}
\usepackage{amssymb,amsmath,amsthm, amsfonts}
\usepackage{multirow}
\usepackage[normalem]{ulem}
\useunder{\uline}{\ul}{}
\usepackage{color}
\usepackage{rotating}
\usepackage{tabularray}
\usepackage{subcaption}

 \usepackage{mathrsfs}
\usepackage{caption}
\usepackage{capt-of}
\usepackage{lipsum}
\usepackage{multibib}
\newcites{app}{References}
\usepackage{txfonts}
\usepackage{pifont}

\definecolor{c1}{HTML}{765D97} 
\definecolor{c2}{HTML}{006400}
\definecolor{c3}{HTML}{fc6160}
\definecolor{myblue}{HTML}{E6F3FC} 
\definecolor{mygray}{HTML}{DBE2E9} 
\definecolor{mygreen}{HTML}{006400} 

\hypersetup{
    colorlinks=true,
    linkcolor=black,
    urlcolor=c1,
    citecolor=c2,
}

\newcommand{\R}[0]{\mathbb{R}}

\title{VFIMamba: Video Frame Interpolation \\ with State Space Models}

\author{%
  \hspace{-5mm} Guozhen Zhang$^{1,2}$\thanks{Work is done during internship at Tencent PCG.~~\textsuperscript{\dag}Corresponding author  (lmwang@nju.edu.cn).} \quad Chunxu Liu$^1$ \quad Yutao Cui$^2$ \quad Xiaotong Zhao$^2$ \quad Kai Ma$^2$ \quad Limin Wang$^{1,3\dagger}$ \\
  $^1$State Key Laboratory for Novel Software Technology, Nanjing University\\
  $^2$Platform and Content Group (PCG), Tencent \quad $^3$Shanghai AI Lab\\
  \newline
  \\
  \newline
  \textbf{\url{https://github.com/MCG-NJU/VFIMamba}}\\
}

\begin{document}

\maketitle

\begin{abstract}
  Inter-frame modeling is pivotal in generating intermediate frames for video frame interpolation (VFI). Current approaches predominantly rely on convolution or attention-based models, which often either lack sufficient receptive fields or entail significant computational overheads. Recently, Selective State Space Models (S6) have emerged, tailored specifically for long sequence modeling, offering both linear complexity and data-dependent modeling capabilities. In this paper, we propose VFIMamba, a novel frame interpolation method for efficient and dynamic inter-frame modeling by harnessing the S6 model. Our approach introduces the Mixed-SSM Block (MSB), which initially rearranges tokens from adjacent frames in an interleaved fashion and subsequently applies multi-directional S6 modeling. This design facilitates the efficient transmission of information across frames while upholding linear complexity. Furthermore, we introduce a novel curriculum learning strategy that progressively cultivates proficiency in modeling inter-frame dynamics across varying motion magnitudes, fully unleashing the potential of the S6 model. Experimental findings showcase that our method attains state-of-the-art performance across diverse benchmarks, particularly excelling in high-resolution scenarios. In particular, on the X-TEST dataset, VFIMamba demonstrates a noteworthy improvement of \textbf{0.80} dB for 4K frames and \textbf{0.96} dB for 2K frames.
\end{abstract}

\begin{figure*}[ht]
    \vspace{-4mm}
    \centering
    \includegraphics[width=0.86\linewidth, trim=5cm 2cm 5.3cm 3cm, clip=true]{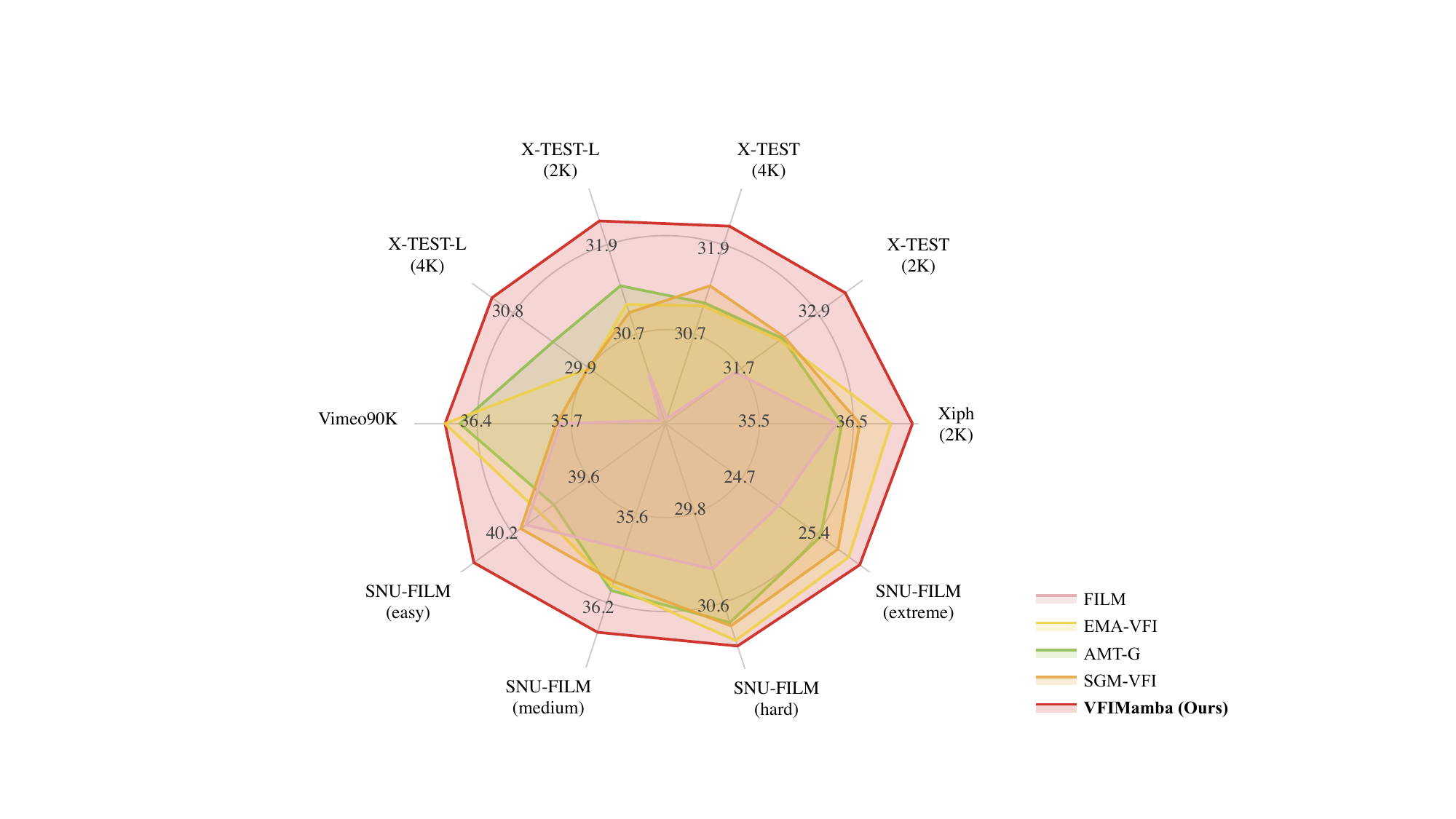}
    \caption{Equipped with the S6 model, our VFIMamba achieves the state-of-the-art performance on benchmarks across different input resolutions. }
    \label{fig:introfig}
    \vspace{-2mm}
\end{figure*}

\input{Sections/01_intro}

\input{Sections/02_related_works}

\input{Sections/03_method}

\input{Sections/04_Experiments}

\section{Conclusion}
In this paper, we have introduced VFIMamba, the first approach to adapt the SSM model to the video frame interpolation task. We devise the Mixed-SSM Block (MSB) for efficient inter-frame modeling using S6. We also explore various rearrangement methods to convert two frames into a sequence, discovering that interleaved rearrangement is more suitable for VFI tasks. Additionally, we propose a curriculum learning strategy to further leverage the potential of the S6 model. Experimental results demonstrate that VFIMamba achieves the state-of-the-art performance across various datasets, in particular highlighting the potential of the SSM model for VFI tasks with high resolution.


\bibliographystyle{icml2024}

\newpage

\section*{Acknowledgement}
This work is supported by the National Key R$\&$D Program of China (No. 2022ZD0160900), the National Natural Science Foundation of China (No. 62076119), the Fundamental Research Funds for the Central Universities (No. 020214380119), the Nanjing University- China Mobile Communications Group Co.,Ltd. Joint Institute, and the Collaborative Innovation Center of Novel Software Technology and Industrialization.

\bibliography{main}

\newpage

\input{Sections/X_supp.tex}


\end{document}

%% file: Sections/01_intro.tex
\section{Introduction}

Video Frame Interpolation (VFI), a fundamental task in video data processing, is gaining substantial attention for its ability to generate intermediate frames between consecutive frames~\citep{liu2017dvf}. Its utility spans many practical applications, including creating slow-motion videos through temporal upsampling~\citep{jiang2018super}, enhancing video refresh rates~\citep{reda2022film}, and generating novel views~\citep{flynn2016deepstereo,szeliski1999prediction}. VFI workflows typically encompass two primary stages~\citep{zhang2023ema}: firstly, capturing the inter-frame dynamics of input consecutive frames; and secondly, leveraging this information to estimate inter-frame motion and generate intermediate frame appearance. In practice, VFI often deals with high-resolution inputs (e.g., 4K)~\citep{sim2021xvfi}, which results in significant object displacement and imposes high demands on the large receptive field of the modules that model information between frames. Additionally, since VFI is commonly applied to long-duration videos such as movies, model speed is also of paramount importance. Thus, \textit{striking a delicate balance between a sufficient receptive field and fast processing speed in modeling inter-frame information} is the key aspect in crafting effective VFI models.

Current methods for modeling inter-frame information predominantly rely on convolutional neural networks (CNNs)~\citep{liu2017dvf,kong2022ifrnet,huang2022rife} and attention-based models~\citep{lu2022vfiformer,zhang2023ema,park2023biformer,liu2024sgm}. However, as illustrated in the first three rows of Table~\ref{tab:intro}, these methods either (1) lack flexibility and cannot adaptively model based on the input, (2) do not have sufficient receptive fields to capture inter-frame correlations at high resolutions, or (3) suffer from prohibitive computational complexity.

On the other hand, Natural Language Processing (NLP) has recently witnessed the emergence of structured state space models (SSMs)~\citep{gu2021s4}. Theoretically, SSMs combine the benefits of Recurrent Neural Networks (RNNs) and CNNs, leveraging the global receptive field characteristic of RNNs and the computational efficiency inherent in CNNs.
One particularly notable SSM is the Selective State Space Model (S6), also known as Mamba~\citep{gu2023mamba}, which has garnered significant attention within the vision community. Mamba's novel feature of making SSM parameters time-variant (i.e., data-dependent) enables it to effectively select relevant context within sequences, a crucial factor for enhancing model performance. However, to the best of our knowledge, \textit{S6 has not yet been applied to low-level video tasks}.

To address the challenges faced by current VFI models and to explore the potential of the S6 model~\citep{gu2023mamba} in low-level video tasks, we propose VFIMamba, a novel frame interpolation method that adapts the S6 model for efficient and dynamic inter-frame modeling. As shown in Table~\ref{tab:intro}, VFIMamba provides the advantages of a global receptive field with linear complexity while maintaining data-dependent adaptability.

Specifically, we introduce the Mixed-SSM Block (MSB) to replace existing modules for inter-frame information transfer. The original S6 model can only process a single sequence, so it is necessary to merge tokens from two frames into one sequence for effective inter-frame modeling. After thorough analysis and exploration, we figured out that interleaving tokens from both frames into a ``super image'' is more suitable for VFI. We then conduct multi-directional SSMs on this image to model inter-frame information. This interleaved approach facilitates interactions between adjacent tokens from different frames during sequence modeling and ensures that the intermediate tokens of any pair of tokens in the sequence are from their spatiotemporal neighborhood. By stacking MSB modules, our model effectively handles complex inter-frame information exchange. Finally, we use the extracted inter-frame features to estimate motion and generate the appearance of intermediate frames.

While the S6 model boasts the advantages listed in Table~\ref{tab:intro}, it is crucial to employ appropriate training strategies to fully exploit its potential. Inspired by \citet{bengio2009curriculum}, we propose a novel curriculum learning strategy that progressively teaches the model to handle inter-frame modeling across varying motion amplitudes. Specifically, while maintaining training on Vimeo90K~\citep{xue2019vimeo}, we incrementally introduce large motion data from X-TRAIN~\citep{sim2021xvfi}, increasing the motion amplitude over time. This learning strategy enables VFIMamba to perform well across a wide range of motion amplitudes, thereby fully unleashing the potential of the S6 model.

To validate the effectiveness of VFIMamba across various types of video data, we conduct extensive testing on different benchmarks. As shown in Figure~\ref{fig:introfig}, VFIMamba achieves the state-of-the-art (SOTA) performance across diverse datasets. This is particularly evident in high-resolution and large-motion datasets such as X-TEST~\citep{sim2021xvfi} and SNU-FILM~\citep{choi2020channel}.

\textbf{Contribution.} In summary, the contributions of this paper are as follows: (1) We are the first to adapt the S6 model to the video frame interpolation task. To better adapt the model for this task, we introduce the Mixed-SSM Block (MSB), providing a solid foundation for future architectural exploration in frame interpolation. (2) We propose a novel curriculum learning strategy that incrementally introduces data with varying motion amplitudes, thereby fully harnessing the potential of the S6 model. (3) Our model achieves the state-of-the-art performance across a wide range of datasets, potentially sparking interest in the exploration of the S6 model within the video low-level community.

\begin{table*}
\centering
\caption{Comparison of the model design for inter-frame modeling of VFIMamba and existing methods. VFIMamba enjoys both the advantages of a large receptive field and linear complexity.}
\label{tab:intro}
\resizebox{0.85\linewidth}{!}{
\begin{tabular}{lcccc} 
\toprule
\textbf{Model}  & \textbf{Data-dependent} & \textbf{Linear complexity} & \textbf{Global receptive field} & \textbf{Representative method} \\ 
\midrule
CNN & \ding{55} & \ding{51}  & \ding{55} &  RIFE~\citep{huang2022rife} \\
Attention  & \ding{51}  & \ding{55}  & \ding{51} &  SGM-VFI~\citep{liu2024sgm} \\
Local Attention  & \ding{51}  & \ding{51}  & \ding{55} & EMA-VFI~\citep{zhang2023ema} \\
\rowcolor[rgb]{0.929,0.902,0.973} Mamba & \ding{51}  & \ding{51}   & \ding{51} & VFIMamba (our work)\\
\midrule
\end{tabular}
}
\end{table*}

%% file: Sections/02_related_works.tex
\section{Related work}

\subsection{Video frame interpolation}
The performance of VFI methods has seen significant advancements with the emergence of deep learning models. \textbf{(1)} CNNs-based approaches~\citep{bao2019depth,liu2017dvf,huang2022rife,niklaus2018context,choi2020channel,zhu2024dual,jia2022ncm,niklaus2017sepconv,kalluri2023flavr}: Initially, DVF~\citep{liu2017dvf} utilized a U-Net-like~\citep{ronneberger2015unet} network to model two input frames and predicted the voxel flow for warping the two frames into the intermediate frame. Following this, CtxSyn~\citep{niklaus2018context} introduced ContextNet and RefineNet, where ContextNet extracts context information from each frame, and RefineNet further refines the coarse intermediate frame produced by warping. RIFE~\citep{huang2022rife} proposed a novel, efficient framework that employs self-distillation to significantly reduce computational load and parameters while maintaining high performance. Due to its simplicity, many convolutional modeling works~\citep{kong2022ifrnet,jia2022ncm} have improved upon RIFE. \textbf{(2)} Attention-based approaches~\citep{lu2022vfiformer,zhang2023ema,park2023biformer,liu2024sgm}: VFIFormer~\citep{lu2022vfiformer} was the first to use attention to model inter-frame information, replacing the encoder part of U-Net with Transformer blocks. After that, EMA-VFI~\citep{zhang2023ema} uses Swin-based~\citep{liu2021swin} local attention to simultaneously capture local appearance and motion information. AMT~\citep{li2023amt} used a multi-scale cost-volume construction similar to RAFT~\citep{teed2020raft} to further enhance motion modeling capabilities. BiFormer~\citep{park2023biformer} introduced quasi-global bilateral attention to further increase the receptive field for large motions. SGM-VFI~\citep{liu2024sgm} introduced sparse global matching to model motion between frames. However, current models struggle to balance sufficient receptive fields with computational overhead. In contrast, our method introduces the first interpolation model based on State Space Models (SSMs)~\citep{gu2023mamba} and further pushes the performance boundaries of VFI tasks.

\subsection{State space models}
In the field of NLP, SSMs~\citep{gu2021s4,smith2022s5,mehta2022gss,fu2022h5} have recently emerged as one of the most promising contenders to challenge the dominance of Transformers. The Structured State Space Sequence Model (S4)~\citep{gu2021s4} was initially introduced for linear complexity modeling of long sequences. Subsequent works have improved its computational efficiency and model capacity. S5~\citep{smith2022s5} proposed a parallel scan and MIMO SSM, and GSS~\citep{mehta2022gss} enhanced the model’s capability by introducing gated mechanisms. Mamba (S6)~\citep{gu2023mamba} has recently stood out due to its data-dependent parameter generation and efficient hardware implementation, outperforming Transformers in long-sequence NLP tasks. In the visual domain, Vim~\citep{zhu2024vim} was the first to permute 2D images into sequences for global modeling using bidirectional SSMs. Vmamba~\citep{liu2024vmamba} extended to four directions and introduced a hierarchical structural design. VideoMamba~\citep{li2024videomamba} was the first to apply S6 in the video domain by permuting all frames into a spatiotemporal sequence. MambaIR~\citep{guo2024mambair} was the first to use the S6 model for image restoration tasks, achieving superior performance over Transformers. In this work, we explore the potential of the S6 model in VFI tasks, validating its effectiveness through detailed analysis and experimentation.

\begin{figure*}[t]
    \vspace{-4mm}
    \centering
    \includegraphics[width=\linewidth, trim=6cm 0cm 6cm 0cm, clip=true]{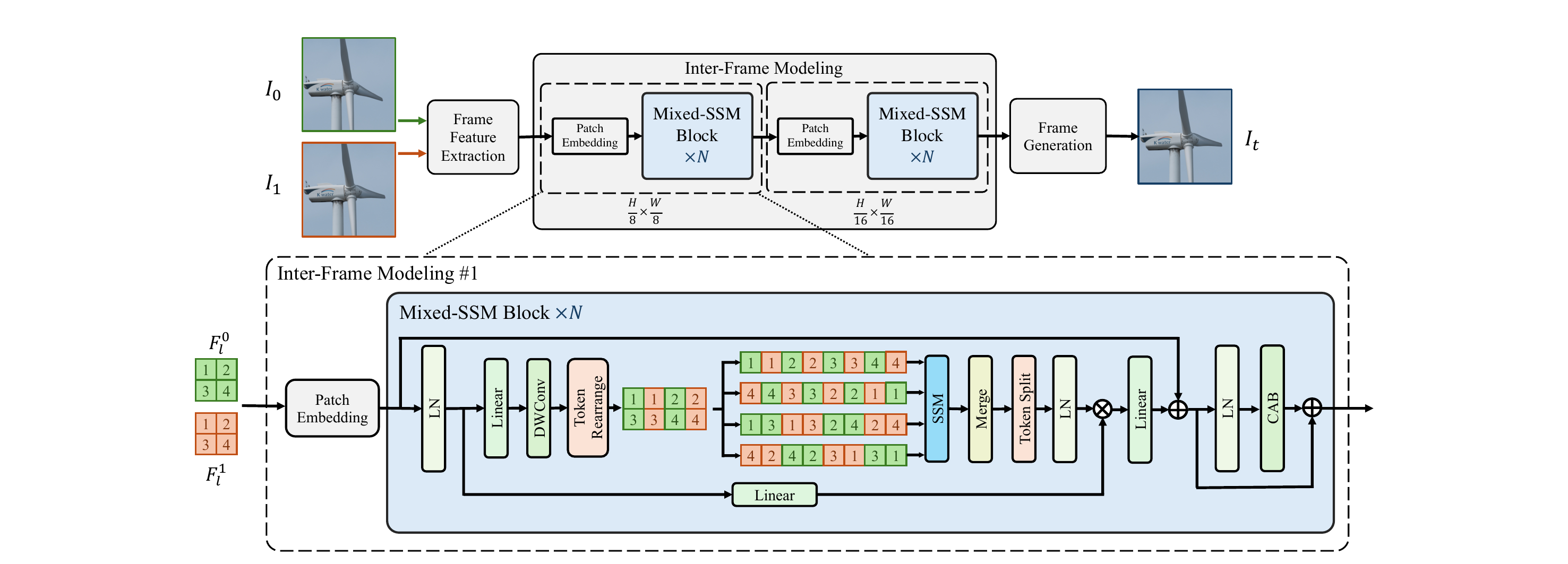}
    \caption{Overall pipeline of VFIMamba. Firstly, a lightweight feature extractor is employed to encode the input frames into shallow features. Subsequently, we utilize the Mixed-SSM Block (MSB) to conduct inter-frame modeling using S6, iterating $N$ times at each scale. Finally, these inter-frame features are leveraged to generate the intermediate frame.}
    \label{fig:method_1}
    \vspace{-2mm}
\end{figure*}

%% file: Sections/03_method.tex
\section{Method}

\subsection{Preliminaries}
\label{sec:3.1}

SSMs are mainly inspired by the continuous linear time-invariant (LTI) systems, which apply an implicit latent state $h(t)\in \R^N$ to map a 1-dimensional sequence or function $x(t)\in \R \rightarrow y(t) \in \R$. Specifically, SSMs can be formulated as an ordinary differential equation (ODE):
\begin{align}
    &h'(t) = A h(t) + B x(t), \\
    & y(t) = C h(t),
\end{align}
where contains evolution matrix $A\in \R^{N\times N}$, projection parameters $B \in \R^{N\times 1}$ and $C\in \R^{1 \times N}$. However, it is hard to solve the above differential equation in deep learning settings and needs to be approximated through discretization. Recent SSMs \citep{gu2021s4} propose to introduce a timescale parameter $\Delta$ to transform the $A,B$ to their discrete counterparts $\bar{A},\bar{B}$, i.e.,
\begin{align}
    &h_t = \bar{A}h_{t-1} + \bar{B}x_t,\\ 
    &y_t = Ch_t, \\ \label{eq:d2}
    &\bar{A} = \exp\left( \Delta A \right),\\ \label{eq:d1}
    &\bar{B} = \left( \Delta A \right)^{-1} \left( \exp \left( \Delta A - I \right) \right)  \cdot \Delta B. 
\end{align}
The above SSMs are performed for each channel separately and their parameters are data-independent, meaning that $\bar{A}$, $\bar{B}$ and ${C}$ are the same for any input in the same channel, limiting their flexibility in sequence modeling. Mamba~\citep{gu2023mamba} proposes the selective SSMs (S6), which dynamically generate the parameters for each input data $x_i \in \R^L$ using the entire $x_i$: 
\begin{equation}
    B_i = S_B x_i, \qquad C_i = S_C x_i, \qquad \Delta_i = \texttt{Softplus}\left(S_{\Delta} x_i\right),
\end{equation}
where $S_B \in \R^{N\times L}, S_C \in \R^{N\times L}, S_{\Delta} \in \R^{L\times L}$ are linear projection layers. The $B_i$ and $C_i$ are shared for all channels of $x_i$, $\Delta_i$ contains $\Delta$ of $L$ channels, and $A$ are the same as previous SSMs. 
By the discretization in equations \ref{eq:d2} and \ref{eq:d1}, $\bar{A}$ and $\bar{B}$ become different based on input data.

\subsection{Overall pipeline}
\label{sec:3.2}

Given two consecutive frames $I_0, I_1 \in \R^{H\times W \times 3}$ along with a timestep $t$, the objective of the frame interpolation task is to generate the intermediate frame $I_t \in \R^{H\times W \times 3}$. As illustrated in Figure~\ref{fig:method_1}, the overall pipeline of VFIMamba consists of three main components: frame feature extraction, inter-frame modeling, and frame generation. Firstly, we employ a set of lightweight convolutional layers to independently extract shallow features from each frame, progressively reducing the resolution to facilitate more efficient inter-frame modeling. This process can be formulated as:
\begin{equation}
F_l^i = \mathcal{L}(I_i),
\end{equation}
where $\mathcal{L}$ represents the set of convolutional layers, and $F_l^i$ denotes the extracted low-level feature for $I_i$. Next, we perform multi-resolution inter-frame modeling using the proposed Mixed-SSM Block (MSB). Each scale comprises $N$ MSBs, and downsampling between scales is achieved through overlapping patch embedding~\citep{wang2022pvt}. We define the resulting inter-frame features as $F_{ssm}^i$. Finally, we utilize these high-quality inter-frame features for frame generation, which involves motion estimation between two frames and appearance refinement:
\begin{equation}
I_t = \mathcal{G}(F_{ssm}^0, F_{ssm}^1),
\end{equation}
where $\mathcal{G}$ denotes the frame generation network. Since this work primarily focuses on exploring the use of SSMs for inter-frame modeling, we largely follow the design from \citet{zhang2023ema} and \citet{huang2022rife} for the frame generation components, with detailed specifications provided in the appendix.

\subsection{State space models for inter-frame modeling}
\label{sec:3.3}

Effective inter-frame modeling is crucial for frame interpolation tasks~\citep{zhang2023ema}. Methods such as RIFE~\citep{huang2022rife} and EMA-VFI~\citep{zhang2023ema} employ simple convolution layers or local attention for inter-frame modeling, achieving high inference speeds but limiting receptive field. Conversely, SGM-VFI~\citep{liu2024sgm} uses global inter-frame attention for motion estimation, which improves performance but sacrifices efficiency. In this work, we propose to use state space models (SSMs), specifically S6~\citep{gu2023mamba}, to achieve both efficiency and effectiveness in inter-frame modeling.

\subsubsection{Mixed-SSM block}
\begin{figure*}[t]
    \centering
    \includegraphics[width=0.88\linewidth, trim=0cm 1.5cm 2cm 1cm, clip=true]{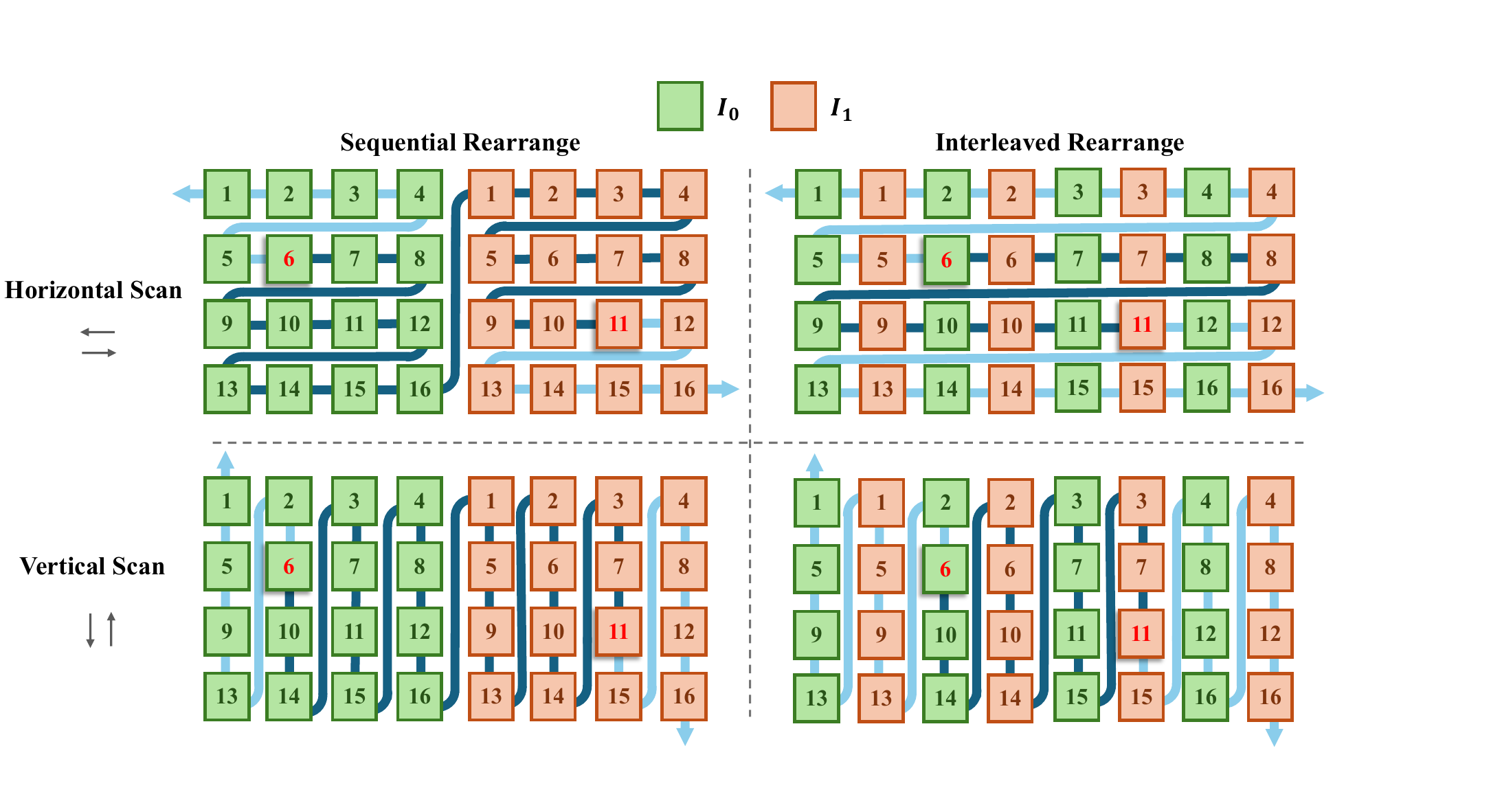}
    \caption{
    Visualizations of different rearrangement methods and scan directions. The choice of rearrangement strategy impacts the information flow during inter-frame modeling with S6. For example, consider the 6-th token from $I_0$ and the 11-th token from $I_1$. In sequential rearrangement, the intermediate tokens introduce too many irrelevant tokens, whereas interleaved rearrangement more effectively preserves the spatiotemporal locality.}
    \label{fig:method_2}
    \vspace{-2mm}
\end{figure*}

As illustrated in Figure~\ref{fig:method_1}, we introduce the Mixed-SSM Block (MSB) for inter-frame modeling. The overall design of the MSB is analogous to Transformers~\citep{vaswani2017attention}, but with two pivotal distinctions: (1) We substitute the attention mechanism with an enhanced S6 Block~\citep{gu2023mamba}, which facilitates global inter-frame modeling with linear complexity. (2) Drawing inspiration from \citet{guo2024mambair} and \citet{behrouz2024mambamixer}, which identified the lack of locality and inter-channel interaction in SSMs, we replace the multilayer perceptron (MLP) with a Channel-Attention Block (CAB)~\citep{hu2018squeeze}.

The original S6 Block is limited to processing one-dimensional sequences, necessitating a strategy for scanning the feature maps of two input frames for inter-frame modeling. As depicted in Figure~\ref{fig:method_2}, there are two primary methods to rearrange the two frames: \textbf{sequential rearrangement}, where the frames are concatenated into a single super image, and \textbf{interleaved rearrangement}, where the tokens of the two frames are interleaved to form a super image. Regardless of the rearrangement method, following \citet{liu2024vmamba}, the super image can be scanned in four directions: horizontally, vertically, and in their respective reverse directions. The S6 Block is then employed to model each direction independently, and the resulting sequences are rearranged and merged back.

\subsubsection{Discussion on how to rearrange}
\label{sec:dis}
Here, we discuss which rearrangement method is better for inter-frame modeling in the context of frame interpolation. First, let us introduce a conclusion from~\citep{ali2024hidden}: the S6 layers can be approximated as hidden attention layers, with the attention weights given by:
\begin{equation}
  \alpha_{i,j} \approx Q_i K_j H_{i,j},
\end{equation}
where
\begin{equation}
Q_i = S_C x_i,\quad K_j = \texttt{ReLU}\left(S_\Delta x_j\right) S_B x_j, \quad H_{i,j} = \exp(\sum_{\substack{k\in (i,j) \\ S_{\Delta} x_k > 0}} (S_{\Delta} x_k)) A, 
\end{equation}
In this formulation, $\alpha_{i,j}$ represents the hidden attention weight of the $j$-th token $x_j$ to the $i$-th token $x_i$ in the sequence. Unlike attention, which calculates weights based solely on the information from tokens $x_i$ ($Q_i$) and $x_j$ ($K_j$), the S6 model includes $H_{i,j}$, which encompasses the contextual information between the $i$-th and $j$-th tokens in the sequence. Based on this conclusion, we observe that in the interleaved rearrangement, the intermediate tokens of any pair of tokens in the sequence are from their spatiotemporal neighborhood. This means that $H_{i,j}$ incorporates more local modeling, which is beneficial for low-level tasks like frame interpolation. Additionally, the number of intermediate tokens between spatiotemporally adjacent tokens is generally smaller in the interleaved rearrangement. In contrast, in the sequential rearrangement, even spatiotemporally adjacent tokens are separated by many unrelated tokens in the sequence. This can introduce noise and interfere with the modeling of the relationship between these tokens. A specific example can be seen in Figure~\ref{fig:method_2}, where the tokens between the 6-th token of the first frame and the 11-th token of the second frame differs significantly between the two rearrangement methods. In summary, we believe that for video frame interpolation, the interleaved rearrangement method is more suitable for better local spatially-aware processing. Our experiments, detailed in Section~\ref{sec:arr}, further validate this conclusion.

\subsection{Curriculum learning for VFIMamba}
\label{sec:3.4}

Despite the advantageous characteristics of the S6 model, such as data dependence and global receptive field, it is crucial to fully exploit its potential through appropriate training strategies. Currently, two main training strategies are employed for frame interpolation: (1) \textbf{Vimeo90K Only}: Most methods training models exclusively on the Vimeo90K~\citep{xue2019vimeo}. Although Vimeo90K offers a rich variety of video content, as analyzed by \citet{liu2024sgm} and \citet{kiefhaber2024benchmarking}, its motions have limited magnitude. This restriction hampers the model's performance on inputs with large motions or high resolution. (2) \textbf{Sequential Learning}: To mitigate the limitations of training solely on Vimeo90K, some methods~\citep{liu2024sgm,park2023biformer} further train the model on X-TRAIN~\citep{sim2021xvfi}, a dataset characterized by large motions and high-resolution content, after initial training on Vimeo90K. While this approach successfully enhances the model's performance on high-resolution data, it often leads to the forgetting of the small-motion modeling capabilities acquired from Vimeo90K.

To address these issues and fully exploit the potential of the S6 model, inspired by \citet{bengio2009curriculum}, we propose a \textbf{curriculum learning} strategy for learning inter-frame modeling capabilities across varying motion magnitudes while maintaining the ability to model small motions. Specifically, while continuing training on Vimeo90K, we progressively incorporated data from X-TRAIN. The original size of X-TRAIN is $512\times 512$, to co-train with Vimeo90K, we first resize the frames to $S\times S$ and then random crop to the same as Vimeo90K. Every $T$ epochs, the resized size $S$ is increased by 10\% (starting from 256), and the temporal interval between selected frames is doubled (starting from 2), which means the motion magnitude increases as training progresses. This strategy enables the model to gradually learn inter-frame modeling capabilities across varying motion magnitudes, starting with smaller motions and progressing to larger ones.

%% file: Sections/04_Experiments.tex
\section{Experiments}

We provide two models: a lightweight model, VFIMamba-S, and a high-performance model, VFIMamba. Both models have $N=3$; the only difference is that VFIMamba has twice the number of channels as VFIMamba-S. As described in Section 3.4, we employ a curriculum learning strategy in which $T=50$ and train for 300 epochs in total. More model configurations and training details are provided in the appendix.

\input{Sections/04_Table2}

\input{Sections/04_Table3}

\begin{figure*}[t]
    \centering
    \includegraphics[width=\linewidth, trim=1.5cm 3.5cm 2cm 3cm, clip=true]{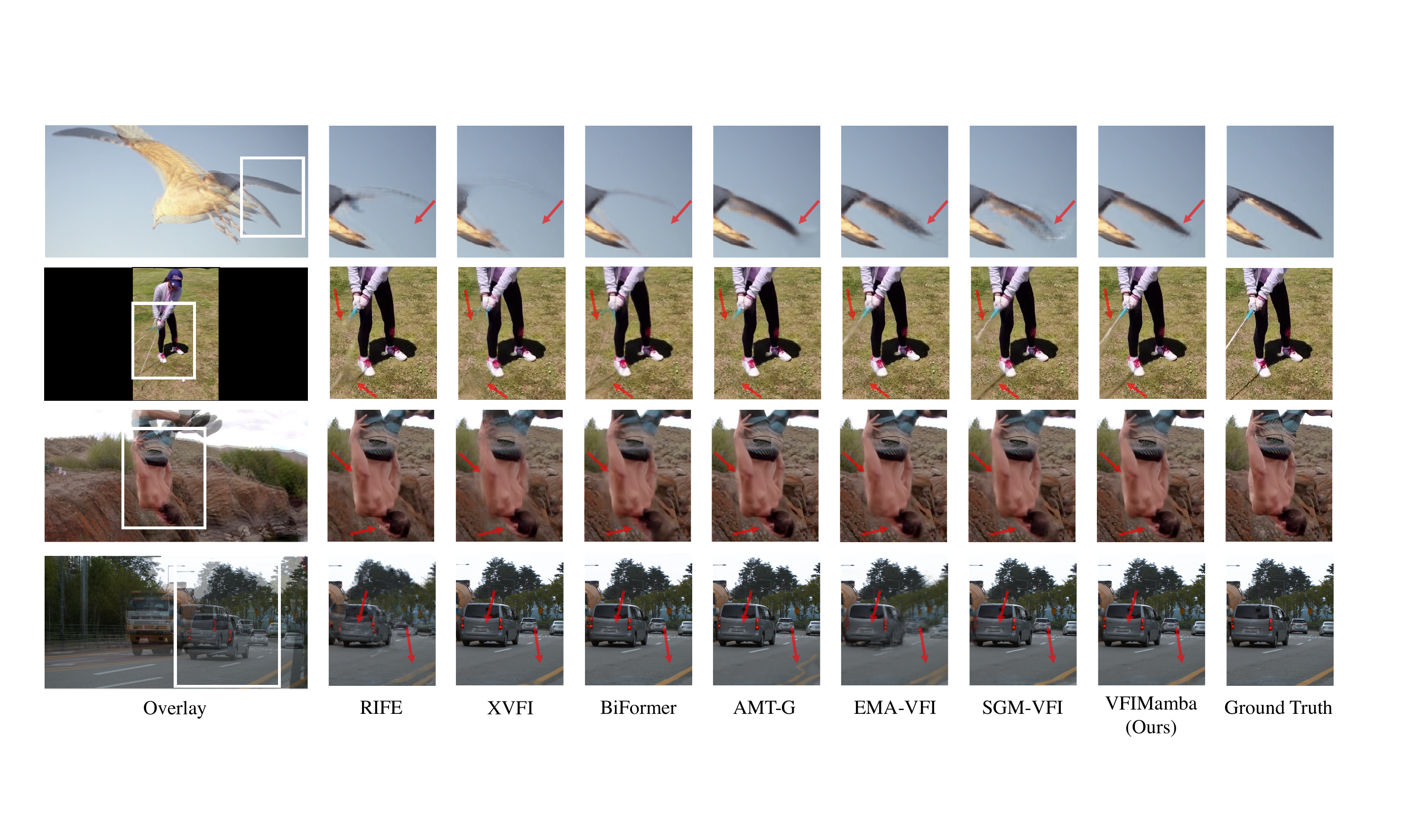}
    \caption{Visualizations from SNU-FILM~\citep{reda2022film} and X-TEST~\citep{sim2021xvfi}. }
    \label{fig:vis}
    \vspace{-3mm}
\end{figure*}

\begin{figure}[t]
    \centering
    \begin{subfigure}[b]{0.45\linewidth}
        \centering
        \includegraphics[width=\linewidth]{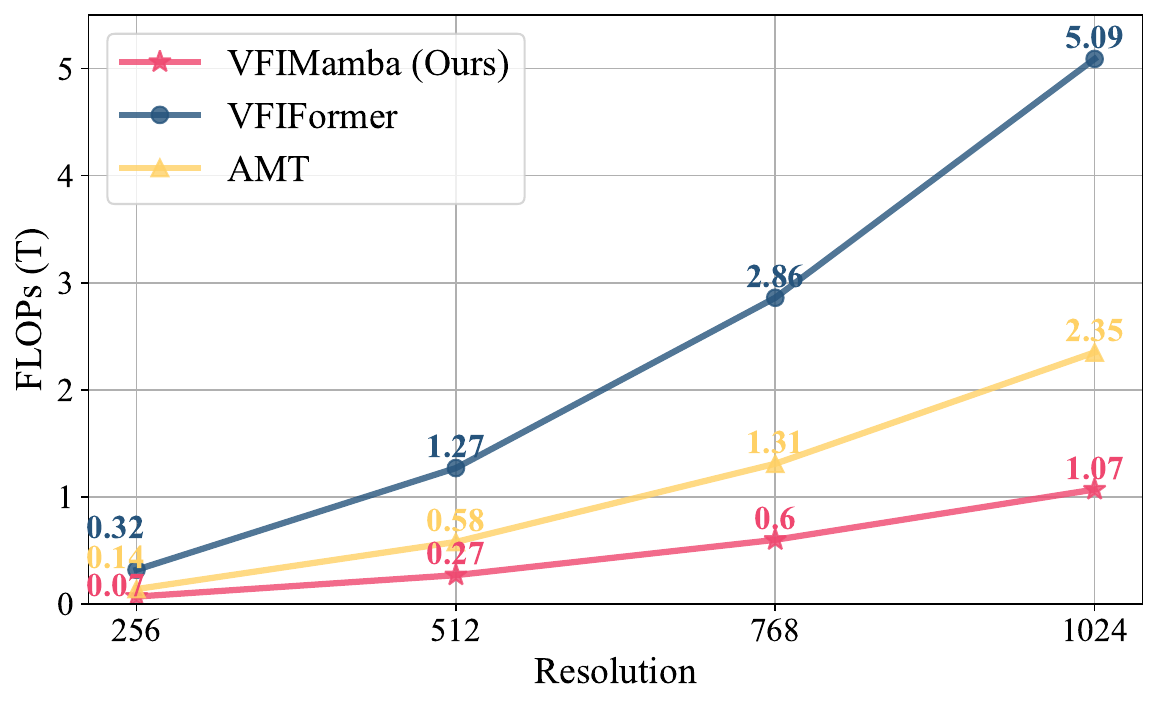}
        \label{fig:c1}
    \end{subfigure}
    \quad
    \begin{subfigure}[b]{0.45\linewidth}
        \centering
        \includegraphics[width=\linewidth]{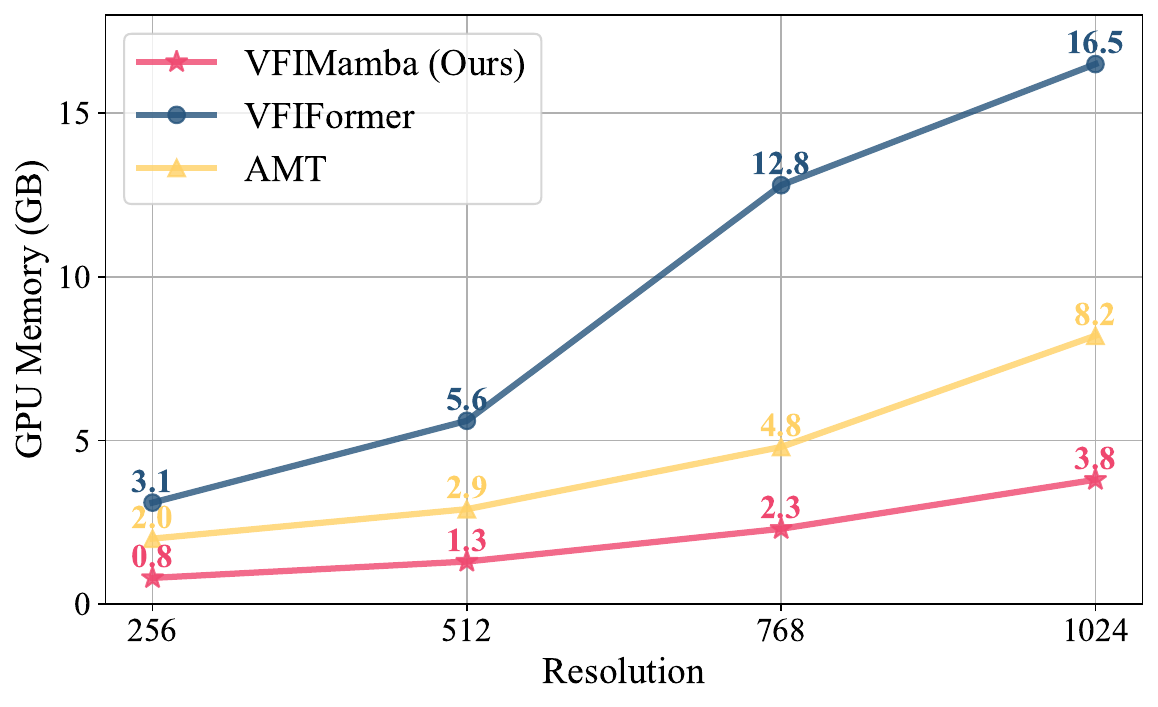}
        \label{fig:c2}
    \end{subfigure}
    \vspace{-0.5cm}
    \caption{Comparisons of FLOPs and GPU memory usage with increasing resolution input.}
    \label{fig:cmps}
    \vspace{-4mm}
\end{figure}

\input{Sections/04_Table4}
\subsection{Comparison with the State-of-the-Art Methods}

\paragraph{Quantitative comparison.} 
To validate the versatility of our proposed VFIMamba, we evaluated its performance (PSNR/SSIM)~\citep{wang2004image} across a variety of well-known benchmarks with different resolutions. The low-resolution datasets include Vimeo90K ($448\times256$)~\citep{xue2019vimeo}, UCF101 ($256\times256$)~\citep{soomro2012ucf101}, and SNU-FILM ($1280\times720$)~\citep{reda2022film}. Notably, SNU-FILM is categorized into four levels of difficulty based on frame intervals: Easy, Medium, Hard, and Extreme. The high-resolution datasets include X-TEST~\citep{sim2021xvfi}, X-TEST-L (a more challenging subset selected by \citet{liu2024sgm}), and Xiph~\citep{xphi1994}. Originally, these datasets are in 4K resolution, and following \citet{zhang2023ema}, we also resize them to 2K for testing.

 For 8x interpolation, we followed the testing procedure of FILM~\citep{reda2022film} and used an iterative approach for frame interpolation. Specifically, we first generated an intermediate frame based on the input two frames, and then, using a divide-and-conquer strategy, we further divided the first frame and the generated intermediate frame, as well as the generated intermediate frame and the last frame, to iteratively generate the remaining frames.

As shown in Tables~\ref{tab:main1} and \ref{tab:main2}, VFIMamba achieves state-of-the-art performance on almost all datasets with FLOPs comparable to efficient models~\citep{kong2022ifrnet,zhang2023ema}. Specifically, in large motion scenarios like X-TEST and X-TEST-L, VFIMamba demonstrates a noteworthy improvement compared with previous metod. This excellent performance underscores the potential of the S6 model in frame interpolation tasks, and we hope it will draw more attention to the application of SSMs in low-level video tasks.

\paragraph{Qualitative comparison.} To further validate the practical effectiveness of VFIMamba, we also present a visual comparison with other frame interpolation methods. As illustrated in Figure~\ref{fig:vis}, the arrows highlight areas where our method excels. VFIMamba demonstrates superior motion estimation and detail preservation in high-motion scenarios compared to other methods. This further substantiates that the incorporation of the S6 model enhances the performance of inter-frame interpolation tasks.

\paragraph{Efficiency comparison.} To validate the efficiency of VFIMamba, we compared the FLOPs and GPU memory usage required by various high-performance methods (\citet{li2023amt} and \citet{lu2022video}) as the resolution increases. As shown in Figure~\ref{fig:cmps}, VFIMamba requires significantly fewer FLOPs and GPU memory as the input resolution grows, demonstrating the effectiveness of the S6 model in the VFI task.

\input{Sections/04_Table5}

\begin{figure*}[t]
    \vspace{-4mm}
    \centering
    \includegraphics[width=0.95\linewidth, trim=1cm 0cm 0.5cm 0cm, clip=true]{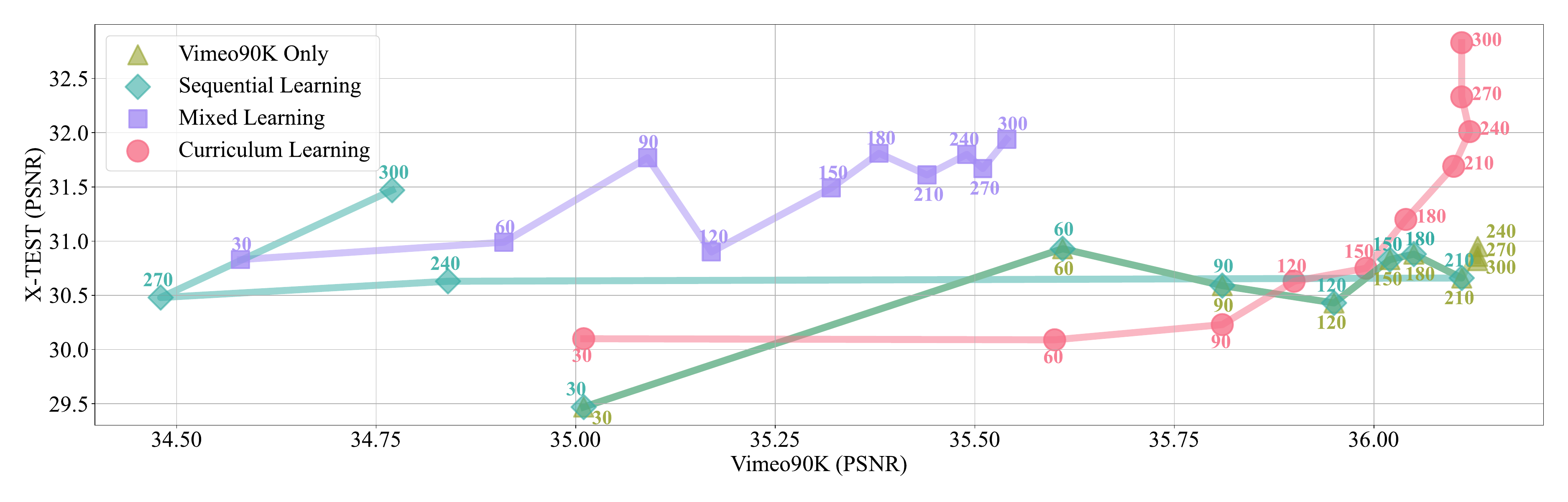}
    \caption{Performance of different learning methods, recorded every 30 epochs. Curriculum learning has the best performance in both the low-resolution and high-resolution benchmarks eventually.}
    \label{fig:curri}
    \vspace{-2mm}
\end{figure*}

\subsection{Ablation Study}

In this section, we conduct ablation studies using the VFIMamba-S model for efficiency.

\paragraph{Effect of the S6 for VFI.} As a core contribution of this work, the S6 model balances computational efficiency and high performance for inter-frame modeling. To validate its effectiveness, as shown in Table~\ref{tab:ab1}, we experimented by removing the SSM model from the MSB (w/o SSM), replacing the MSB with convolutions from RIFE~\citep{huang2022rife} (Convolution), or local inter-frame attention from EMA-VFI~\citep{zhang2023ema} (Local Attention), or global inter-frame attention~\citep{liu2024sgm} (Full Attention). We observed that only removing the S6 model resulted in a parameter reduction of only 0.7M but led to a significant performance drop across various datasets, underscoring the importance of S6. In comparisons with Convolution and Local Attention, we found that although the S6 model is relatively slower due to its multiple scanning directions, it achieves substantial performance improvements. Compared to Full Attention, S6 not only surpasses its performance but also offers faster inference speed and lower memory consumption. In summary, the S6 model indeed achieves a balance between computational efficiency and performance compared to existing models.

\paragraph{Frame rearrangement for inter-frame modeling.}
\label{sec:arr}
The rearrangement of input frames is crucial for inter-frame modeling using the S6 model. As analyzed in Section~\ref{sec:dis}, we posit that interleaved rearrangement is more suitable for VFI tasks, and we provide experimental validation here. As shown in Table~\ref{tab:ab2}, we experimented with two different rearrangement methods in both horizontal and vertical scans. The results demonstrate that using interleaved rearrangement consistently achieves the best performance across all datasets, with significant improvements over other methods. These findings further validate our analysis that interleaved rearrangement offers superior spatiotemporal local modeling capabilities for VFI.

\paragraph{Explore different learning strategy.}
As described in Section~\ref{sec:3.4}, we proposed a curriculum learning strategy to fully harness the global modeling capabilities of the S6 model. In Figure~\ref{fig:curri}, we present the performance of different learning strategies over training epochs on both Vimeo90K and X-TEST. In addition to the Vimeo90K Only and Sequential Learning strategies mentioned in Section~\ref{sec:3.4}, we also compared a baseline approach where the two datasets were directly mixed for training (Mixed Learning). The results indicate that as epochs increase, the Vimeo90K Only strategy improves performance exclusively on Vimeo90K with negligible change on X-TEST. Sequential Learning, while eventually enhancing X-TEST performance, significantly degrades performance on Vimeo90K. Mixed Learning shows a gradual increase in performance on both datasets but fails to achieve competitive results. Our proposed curriculum learning strategy, however, achieves the best performance on both datasets simultaneously by the end of training. 

\input{Sections/X_Table6}

\paragraph{Generalization of curriculum learning}
To validate the generalization capability of curriculum learning, we also trained the RIFE~\citep{huang2022rife} and EMA-VFI~\citep{zhang2023ema} from scratch using curriculum learning. As shown in Table~\ref{tab:currigen}, after training, all models maintained their performance on the low-resolution dataset Vimeo90K while significantly improving performance on the X-TEST and SNU-FILM, fully verified the generalization of curriculum learning. Among these, our VFIMamba achieved the most significant improvement and the highest performance ceiling, further demonstrating the potential of the S6 model.

%% file: Sections/04_Table2.tex
\begin{table}
\centering
\caption{Quantitative comparison with SOTA methods on the low-resolution datasets, in terms of
PSNR/SSIM~\citep{wang2004image}. The best results and the second best results are \textcolor{red}{\textbf{boldfaced}} and \textcolor[rgb]{0,0.439,0.753}{\uline{underlined}} respectively. FLOPs was calculated for 720p input. $\star$ indicates the results copied from \citet{zhang2023ema} and \citet{li2023amt}. In ``Training Dataset'', ``V'' stands for the triplet of Vimeo-90K and ``X'' stands for X-TRAIN.  In ``Runtime'', we evaluate the inference speed of each method on $1024\times 1024$ resolution inputs by a 2080Ti GPU. In ``Average'', we calculate the average performance of each method in terms of PSNR and SSIM. }
\label{tab:main1}
\vspace{2mm}
\resizebox{\linewidth}{!}{%
\begin{tabular}{cccccccclcc} 
\toprule
\multirow{2}{*}{} & \multirow{2}{*}{\begin{tabular}[c]{@{}c@{}}Training\\Dataset\end{tabular}} & \multirow{2}{*}{Vimeo-90K} & \multirow{2}{*}{UCF101} & \multicolumn{4}{c}{SNU-FILM~\citep{choi2020channel}} & \multicolumn{1}{c}{\multirow{2}{*}{Average}} & \multirow{2}{*}{FLOPs (T)} & \multirow{2}{*}{\begin{tabular}[c]{@{}c@{}}Runtime (ms)\end{tabular}} \\ 
\cmidrule(lr){5-5}\cmidrule(lr){6-6}\cmidrule(lr){7-7}\cmidrule(lr){8-8}
 & & \multicolumn{1}{l}{\citep{xue2019vimeo}}  & \multicolumn{1}{l}{\citep{soomro2012ucf101}}  & easy & medium & hard & extreme & \multicolumn{1}{c}{} &  &  \\ 
\midrule
DAIN$\star$~\citep{bao2019depth} & V & 34.71/0.9756 & 34.99/0.9683 & 39.73/0.9902 & 35.46/0.9780 & 30.17/0.9335 & 25.09/0.8584 & 33.36/0.9507 & 5.51 & 897.8 \\
AdaCof$\star$~\citep{lee2020adacof} & V & 34.47/0.9730 & 34.90/0.9680 & 39.80/0.9900 & 35.05/0.9754 & 29.46/0.9244 & 24.31/0.8439 & 33.00/0.9458 & 0.36 & 85.1 \\
CAIN$\star$~\citep{choi2020channel} & V & 34.65/0.9730 & 34.91/0.9690 & 39.89/0.9900 & 35.61/0.9776 & 29.90/0.9292 & 24.78/0.8507 & 33.29/0.9483 & 1.29 & 102.4 \\
Softsplat~\citep{niklaus2020softmax} & V & 36.13/0.9805 & 35.39/0.9697 & \textcolor[rgb]{0,0.439,0.753}{\uline{40.26}}/0.9911 & 36.07/0.9798 & 30.53/0.9365 & 25.16/0.8604 & 33.92/0.9530 & 0.94 & 266.4 \\
XVFI~\citep{sim2021xvfi} & V & 35.09/0.9759 & 35.17/0.9685 & 39.93/0.9907 & 35.37/0.9782 & 29.58/0.9276 & 24.17/0.8450 & 33.22/0.9477 & 0.37 & 165.2 \\
M2M-VFI~\citep{hu2022many} & V & 35.47/0.9778 & 35.28/0.9694 & 39.66/0.9904 & 35.74/0.9794 & 30.30/0.9360 & 25.08/0.8604 & 33.59/0.9522 & 0.26 & 60.9 \\
RIFE~\citep{huang2022rife} & V & 35.61/0.9779 & 35.28/0.9690 & 39.80/0.9903 & 35.76/0.9787 & 30.36/0.9351 & 25.27/0.8601 & 33.68/0.9519 & 0.20 & 35.2 \\
IFRNet-L~\citep{kong2022ifrnet} & V & 36.20/0.9808 & 35.42/0.9698 & 40.10/0.9906 & 36.12/0.9797 & 30.63/0.9368 & 25.26/0.8609 & 33.96/0.9531 & 0.79 & 115.3 \\
EMA-VFI-S~\citep{zhang2023ema}  & V & 36.07/0.9797 & 35.34/0.9696 & 39.81/0.9906 & 35.88/0.9795 & 30.69/0.9375 & 25.47/0.8632 & 33.88/0.9534 & 0.20 & 76.4 \\
EMA-VFI~\citep{zhang2023ema}  & V & \textcolor{red}{\textbf{36.64}}/\textcolor{red}{\textbf{0.9819}} & \textcolor{red}{\textbf{35.48}}/\textcolor[rgb]{0,0.439,0.753}{\uline{0.9701}} & 39.98/0.9910 & 36.09/\textcolor[rgb]{0,0.439,0.753}{\uline{0.9801}} & \textcolor[rgb]{0,0.439,0.753}{\uline{30.94}}/\textcolor[rgb]{0,0.439,0.753}{\uline{0.9392}} & \textcolor[rgb]{0,0.439,0.753}{\uline{25.69}}/\textcolor[rgb]{0,0.439,0.753}{\uline{0.8661}} & \textcolor[rgb]{0,0.439,0.753}{\uline{34.14}}/\textcolor[rgb]{0,0.439,0.753}{\uline{0.9547}} & 0.91 & 239.6 \\
AMT-L~\citep{li2023amt}  & V & 36.35/0.9815 & 35.39/0.9698 & 39.95/\textcolor{red}{\textbf{0.9913}} & 36.09/\textcolor{red}{\textbf{0.9805}} & 30.75/0.9384 & 25.41/0.8638 & 33.99/0.9542 & 0.58 & 183.42 \\
AMT-G~\citep{li2023amt}  & V & \textcolor[rgb]{0,0.439,0.753}{\uline{36.53}}/\textcolor[rgb]{0,0.439,0.753}{\uline{0.9817}} & 35.41/0.9699 & 39.88/\textcolor{red}{\textbf{0.9913}} & 36.12/\textcolor{red}{\textbf{0.9805}} & 30.78/0.9385 & 25.43/0.8644 & 34.03/0.9544 & 2.07 & 403.7 \\
SGM-VFI~\citep{liu2024sgm} & V+X & 35.81/0.9793 & 35.34/0.9693 & 40.14/0.9907 & 36.06/0.9795 & 30.81/0.9375 & 25.59/0.8646 & 33.96/0.9535 & 1.78 & 942.9 \\ \midrule
VFIMamba-S & V+X & 36.09/0.9800 & 35.36/0.9696 & 40.21/0.9909 & \textcolor[rgb]{0,0.439,0.753}{\uline{36.17}}/0.9800 & 30.80/0.9381 & 25.59/0.8655 & 34.04/0.9540 & 0.24 & 128.0 \\
VFIMamba & V+X & \textcolor{red}{\textbf{36.64}}/\textcolor{red}{\textbf{0.9819}} & \textcolor[rgb]{0,0.439,0.753}{\uline{35.45}}/\textcolor{red}{\textbf{0.9702}} & \textcolor{red}{\textbf{40.51}}/\textcolor[rgb]{0,0.439,0.753}{\uline{0.9912}} & \textcolor{red}{\textbf{36.40}}/\textcolor{red}{\textbf{0.9805}} & \textcolor{red}{\textbf{30.99}}/\textcolor{red}{\textbf{0.9401}} & \textcolor{red}{\textbf{25.79}}/\textcolor{red}{\textbf{0.8682}} & \textcolor{red}{\textbf{34.30}}/\textcolor{red}{\textbf{0.9554}} & 0.94 & 310.9 \\
\bottomrule
\end{tabular}
}
\vspace{-2mm}
\end{table}

%% file: Sections/04_Table3.tex
\begin{table}
\centering
\caption{Quantitative comparison with SOTA methods on high-resolution datasets. ``OOM'' indicates ``Out of Memory'' on V100. All results are obtained through the same evaluation procedure.}
\label{tab:main2}
\vspace{2mm}
\resizebox{0.85\linewidth}{!}{%
\begin{tabular}{ccccccccc} 
\toprule
\multirow{2}{*}{} & \multirow{2}{*}{\begin{tabular}[c]{@{}c@{}}Training\\Dataset\end{tabular}} & \multicolumn{2}{c}{X-TEST~\citep{sim2021xvfi}} & \multicolumn{2}{c}{X-TEST-L~\citep{liu2024sgm}} & \multicolumn{2}{c}{Xiph~\citep{xphi1994}} & \multirow{2}{*}{Average} \\ 
\cmidrule(lr){3-3}\cmidrule(lr){4-4}\cmidrule(lr){5-5}\cmidrule(lr){6-6}\cmidrule(lr){7-7}\cmidrule(lr){8-8}
 &  & 2K & 4K & 2K & 4K & 2K & 4K &  \\ 
\midrule
XVFI~\citep{sim2021xvfi} & X & 31.15/0.9144 & 30.12/0.9045 & 29.82/0.8951 & 29.02/0.8866 & 34.76/0.9258 & 32.84/0.8810 & 31.29/0.9012 \\
M2M-VFI~\citep{hu2022many} & V & 32.13/0.9258 & 30.89/0.9138 & 30.90/0.9092 & 29.73/0.9001 & 36.44/0.9427 & 33.92/0.8992 & 32.34/0.9151 \\
RIFE~\citep{huang2022rife} & V & 31.10/0.8972 & 30.13/0.8927 & 29.87/0.8805 & 28.98/0.8756 & 36.19/0.9380 & 33.76/0.8940 & 31.67/0.8963 \\
FILM~\citep{reda2022film} & V & 31.61/0.9174 & OOM & 30.18/0.8960 & OOM & 36.32/0.9343 & 33.27/0.8760 & / \\
IFRNet-L~\citep{kong2022ifrnet} & V & 31.78/0.9147 & 30.66/0.9050 & 30.76/0.8963 & 29.74/0.8884 & 36.21/0.9374 & 34.25/0.8946 & 32.23/0.9061 \\
FLDR~\citep{Nottebaum:2022:EFE} & X & 31.12/0.9092 & 30.46/0.9041 & 29.90/0.8906 & 29.30/0.8879 & 34.80/0.9280 & 33.00/0.8862 & 31.43/0.9010 \\
BiFormer~\citep{park2023biformer} & V+X & 31.32/0.9200 & 31.32/0.9215 & 30.36/0.9068 & 30.14/0.9069 & 34.20/0.9246 & 33.49/0.8953 & 31.81/0.9125 \\
EMA-VFI-S~\citep{zhang2023ema} & V & 30.91/0.9000 & 29.91/0.8951 & 29.51/0.8775 & 28.60/0.8733 & 36.55/0.9421 & 34.25/0.9020 & 31.62/0.8983 \\
AMT-L~\citep{li2023amt} & V & 32.08/0.9277 & 30.96/0.9147 & 31.09/0.9103 & 30.12/0.9019 & 36.27/0.9402 & 34.49/0.9030 & 32.50/0.9163 \\
AMT-G~\citep{li2023amt} & V & 32.35/0.9300 & 31.12/0.9157 & 31.35/0.9125 & 30.33/0.9036 & 36.38/0.9410 & \textcolor{red}{\textbf{34.63}}/\textcolor[rgb]{0,0.439,0.753}{\uline{0.9039}} & 32.69/0.9178 \\
SGM-VFI~\citep{liu2024sgm} & V+X & 32.38/0.9272 & 31.35/0.9179 & 30.99/0.9072 & 29.91/0.8972 & 36.57/0.9424 & 34.23/0.9021 & 32.57/0.9157 \\ \midrule
VFIMamba-S & V+X & \textcolor[rgb]{0,0.439,0.753}{\uline{32.84}}/\textcolor[rgb]{0,0.439,0.753}{\uline{0.9328}} & \textcolor[rgb]{0,0.439,0.753}{\uline{31.73}}/\textcolor[rgb]{0,0.439,0.753}{\uline{0.9238}} & \textcolor[rgb]{0,0.439,0.753}{\uline{31.58}}/\textcolor[rgb]{0,0.439,0.753}{\uline{0.9169}} & \textcolor[rgb]{0,0.439,0.753}{\uline{30.50}}/\textcolor[rgb]{0,0.439,0.753}{\uline{0.9077}} & \textcolor[rgb]{0,0.439,0.753}{\uline{36.72}}/\textcolor[rgb]{0,0.439,0.753}{\uline{0.9428}} & 34.32/0.9034 & \textcolor[rgb]{0,0.439,0.753}{\uline{32.95}}/\textcolor[rgb]{0,0.439,0.753}{\uline{0.9212}} \\
VFIMamba & V+X & \textcolor{red}{\textbf{33.34}}/\textcolor{red}{\textbf{0.9361}} & \textcolor{red}{\textbf{32.15}}/\textcolor{red}{\textbf{0.9246}} & \textcolor{red}{\textbf{32.22}}/\textcolor{red}{\textbf{0.9259}} & \textcolor{red}{\textbf{31.05}}/\textcolor{red}{\textbf{0.9159}} & \textcolor{red}{\textbf{37.13}}/\textcolor[rgb]{0,0.439,0.753}{\uline{0.9451}} & \textcolor[rgb]{0,0.439,0.753}{\uline{34.62}}/\textcolor{red}{\textbf{0.9059}} & \textcolor{red}{\textbf{33.42}}/\textcolor{red}{\textbf{0.9256}} \\
\bottomrule
\end{tabular}
}
\vspace{-2mm}
\end{table}

%% file: Sections/04_Table4.tex
\begin{table*}[t]
\centering
\caption{Ablation on different models for inter-frame modeling. We use the V100 GPU for evaluating and ``OOM'' indicates ``Out of Memory''.}
\label{tab:ab1}
\setlength{\extrarowheight}{0pt}
\addtolength{\extrarowheight}{\aboverulesep}
\addtolength{\extrarowheight}{\belowrulesep}
\setlength{\aboverulesep}{0pt}
\setlength{\belowrulesep}{0pt}
\resizebox{0.9\linewidth}{!}{%
\begin{tabular}{cccccccc} 
\toprule
\multirow{2}{*}{Model}           & \multirow{2}{*}{Vimeo90K} & \multicolumn{2}{c}{X-TEST}  & \multicolumn{2}{c}{SNU-FILM} & \multirow{2}{*}{Params (M)} & \multirow{2}{*}{\begin{tabular}[c]{@{}c@{}}720p Inference\\Time (ms)\end{tabular}}  \\ 
\cmidrule(lr){3-3}\cmidrule(lr){4-4}\cmidrule(lr){5-5}\cmidrule(lr){6-6}
                                 &                            & 2K           & 4K           & hard         & extreme       &                             &                                                                                      \\ 
\midrule
w/o S6                          & 35.62/0.9771               & 28.94/0.8517 & 27.12/0.8436 & 30.41/0.9341 & 25.14/0.8567  & 16.1                        & \textbf{51}                                                                                   \\
Convolution                      & 35.86/0.9790               & 31.58/0.9167 & 30.24/0.9044 & 30.61/0.9365 & 25.49/0.8631  & 23.4                        & 55                                                                                   \\
Local Attention                  & 35.92/0.9790               & 30.49/0.8917 & 30.00/0.8845 & 30.47/0.9338 & 25.46/0.8625  & \textbf{15.6}                        & 59                                                                                   \\
Full Attention                   & 36.04/0.9798               & OOM          & OOM          & 30.55/0.9367 & 25.35/0.8602  & \textbf{15.6}                        & 336                                                                                  \\ 
\midrule
 S6 & \textbf{36.12/0.9802}               & \textbf{32.84/0.9328} & \textbf{31.73/0.9238} & \textbf{30.80/0.9381} & \textbf{25.59/0.8655}  & 16.8                        & 77                                                                                   \\
\bottomrule
\end{tabular}
}
\vspace{2mm}
\end{table*}

%% file: Sections/04_Table5.tex
\begin{table*}[t]
\centering
\caption{Ablation on different rearrangement approachs. ``Sequential" means sequential rearrangement and ``Interleaved" represents interleaved rearrangement .}
\label{tab:ab2}
\setlength{\extrarowheight}{0pt}
\addtolength{\extrarowheight}{\aboverulesep}
\addtolength{\extrarowheight}{\belowrulesep}
\setlength{\aboverulesep}{0pt}
\setlength{\belowrulesep}{0pt}
\resizebox{0.8\linewidth}{!}{%
\begin{tabular}{ccccccc} 
\toprule
\multirow{2}{*}{Horizontal Scan} & \multirow{2}{*}{Vertical Scan} & \multirow{2}{*}{Vimeo-90K} & \multicolumn{2}{c}{X-TEST} & \multicolumn{2}{c}{SNU-FILM} \\ 
\cmidrule(lr){4-4}\cmidrule(lr){5-5}\cmidrule(lr){6-6}\cmidrule(lr){7-7}
 &  &  & 2K & 4K & hard & extreme \\ 
\midrule
Sequential & Sequential & 35.55/0.9765 & 28.07/0.8327 & 26.75/0.8327 & 30.24/0.9319 & 25.03/0.8545 \\
Sequential & {\cellcolor[rgb]{0.847,0.847,0.847}}Interleaved & 35.76/0.9784 & 31.69/0.9226 & 30.45/0.9078 & 30.32/0.9342 & 25.21/0.8611 \\
{\cellcolor[rgb]{0.847,0.847,0.847}}Interleaved & Sequential & 35.79/0.9785 & 31.49/0.9221 & 30.35/0.9053 & 30.12/0.9331 & 25.11/0.8602 \\
{\cellcolor[rgb]{0.847,0.847,0.847}}Interleaved & {\cellcolor[rgb]{0.847,0.847,0.847}}Interleaved & \textbf{36.12/0.9802} & \textbf{32.84/0.9328} & \textbf{31.73/0.9238} & \textbf{30.80/0.9381} & \textbf{25.59/0.8655} \\
\bottomrule
\end{tabular}
}
\vspace{3mm}
\end{table*}

%% file: Sections/X_Table6.tex
\begin{table}
\centering
\caption{Performance of different methods without or with curriculum learning.}
\label{tab:currigen}
\setlength{\extrarowheight}{0pt}
\addtolength{\extrarowheight}{\aboverulesep}
\addtolength{\extrarowheight}{\belowrulesep}
\setlength{\aboverulesep}{0pt}
\setlength{\belowrulesep}{0pt}
\resizebox{0.85\linewidth}{!}{%
\begin{tabular}{ccccccc} 
\toprule
                            & \multirow{2}{*}{\begin{tabular}[c]{@{}c@{}}Curriculum\\Learning\end{tabular}} & \multirow{2}{*}{Vimeo90K}                                & \multicolumn{2}{c}{X-TEST}                                                                                                              & \multicolumn{2}{c}{SNU-FILM}                                                                                                             \\ 
\cmidrule(lr){4-5}\cmidrule(lr){6-7}
                            &                                                                               &                                                           & 2K                                                                 & 4K                                                                 & hard                                                               & extreme                                                             \\ 
\midrule
\multirow{2}{*}{RIFE}       & \ding{55}                                                                     & 35.61/0.9797                                              & 31.10/0.8972                                                       & 30.13/0.8927                                                       & 30.36/0.9375                                                       & 25.27/0.8601                                                        \\
                            & {\cellcolor[rgb]{0.847,0.847,0.847}}\ding{51}                                 & {\cellcolor[rgb]{0.847,0.847,0.847}}35.60/0.9797          & {\cellcolor[rgb]{0.847,0.847,0.847}}31.40/0.9142                   & {\cellcolor[rgb]{0.847,0.847,0.847}}30.23/0.9011                   & {\cellcolor[rgb]{0.847,0.847,0.847}}30.47/0.9376                   & {\cellcolor[rgb]{0.847,0.847,0.847}}25.38/0.8619                    \\ 
\midrule
\multirow{2}{*}{EMA-VFI-S}  & \ding{55}                                                                     & 36.07/0.9797                                              & 30.91/0.9000                                                       & 29.91/0.8951                                                       & 30.69/0.9375                                                       & 25.47/0.8632                                                        \\
                            & {\cellcolor[rgb]{0.847,0.847,0.847}}\ding{51}                                 & {\cellcolor[rgb]{0.847,0.847,0.847}}36.05/0.9797          & {\cellcolor[rgb]{0.847,0.847,0.847}}31.15/0.9083                   & {\cellcolor[rgb]{0.847,0.847,0.847}}29.98/0.8988                   & {\cellcolor[rgb]{0.847,0.847,0.847}}30.73/0.9379                   & {\cellcolor[rgb]{0.847,0.847,0.847}}25.53/0.8652                    \\ 
\midrule
\multirow{2}{*}{VFIMamba-S} & \ding{55}                                                                     & \textbf{36.13}/\textbf{0.9802}                            & 30.82/0.8997                                                       & 29.87/0.8949                                                       & 30.58/0.9378                                                       & 25.30/0.8620                                                        \\
                            & {\cellcolor[rgb]{0.847,0.847,0.847}}\ding{51}                                 & {\cellcolor[rgb]{0.847,0.847,0.847}}36.12/\textbf{0.9802} & {\cellcolor[rgb]{0.847,0.847,0.847}}\textbf{32.84}/\textbf{0.9328} & {\cellcolor[rgb]{0.847,0.847,0.847}}\textbf{31.73}/\textbf{0.9238} & {\cellcolor[rgb]{0.847,0.847,0.847}}\textbf{30.80}/\textbf{0.9381} & {\cellcolor[rgb]{0.847,0.847,0.847}}\textbf{25.59}/\textbf{0.8655}  \\
\bottomrule
\end{tabular}
}
\end{table}

%% file: Sections/X_supp.tex
\appendix

\section{Appendix}

\begin{figure*}[h]
    \centering
    \includegraphics[width=0.8\linewidth, trim=4cm 4cm 4cm 3.5cm, clip=true]{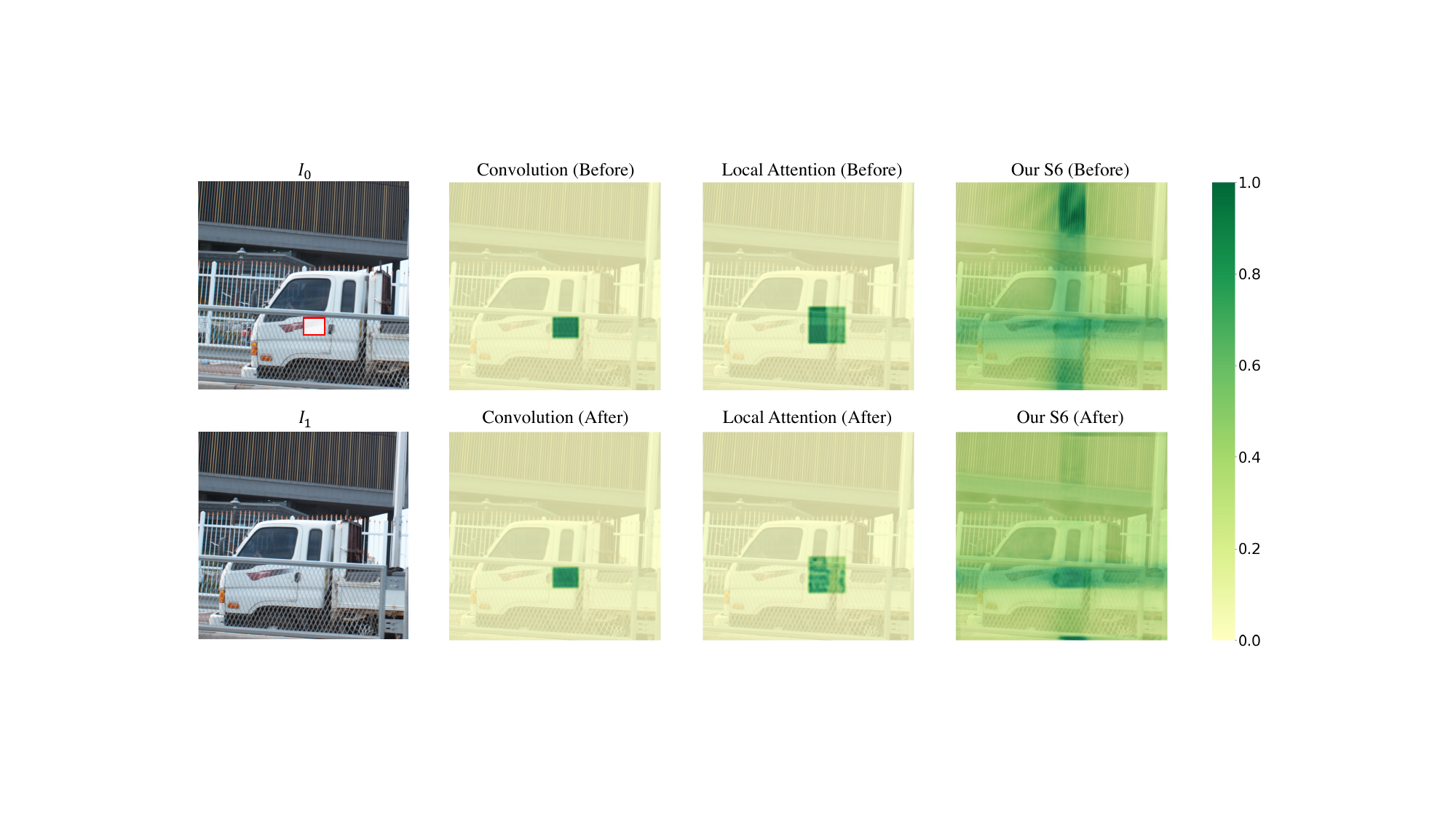}
    \caption{
    Visualizations of Effective Receptive Field (ERF)~\citep{luo2016understanding} of different models before and after training. We utilize the \textcolor{red}{red} area in $I_0$ to inspect its corresponding ERF in $I_1$. S6 model has a significantly larger receptive field and becomes more accurate after training.}
    \label{fig:erf}
    \vspace{-2mm}
\end{figure*}

\subsection{Broader impact}
\label{sm:impact}

In this work, we introduce VFIMamba, the first video frame interpolation model based on SSMs. Video frame interpolation has wide-ranging applications in real-world video data processing, such as increasing the frame rate of AI-generated videos and generating slow-motion videos. Enhancing performance in various scenarios is crucial. However, as a research-oriented work, we trained our model on a very limited set of datasets (Vimeo90K~\citep{xue2019vimeo} and X-TRAIN~\citep{sim2021xvfi}), which might result in some degree of overfitting. Consequently, there could be significant artifacts when applied in real-world usage. This issue can be mitigated by training on a more diverse and extensive set of datasets.

\subsection{Limitations and future work}
\label{sm:limit}

As the first work to explore the application of SSM models in frame interpolation tasks, we have achieved high performance, but there are still some limitations. First, although our method is much faster than attention-based methods, it still falls short of real-time requirements. Future work on designing a more efficient SSMs would be highly valuable. Second, in this work, we primarily focused on the role of SSM in inter-frame modeling and did not explore its use in the frame generation module. In the future, directly using SSM for generating intermediate frames could also be a promising direction for exploration.

\subsection{Visualizations on effective receptive field}

To further evaluate the effective receptive field (ERF) of the S6 model in comparison with other efficient models (CNN, Local Attention) for inter-frame modeling, we used the method described by \citet{luo2016understanding}. Given a specific region in $I_0$, we visualized the corresponding receptive fields in $I_1$ for different methods.

As shown in Figure~\ref{fig:erf}, when the motion between $I_0$ and $I_1$ is significant, neither convolution nor local attention can focus on the corresponding region in  $I_1$ before or after training. In contrast, the S6 model exhibits a larger global receptive field even before training, with notable concentration in both horizontal and vertical directions. We attribute this to the sequence rearrangement, where tokens closer together tend to have higher weights, a phenomenon also observed in VMamba~\citep{liu2024vmamba}.

After training, the S6 model's focus becomes more concentrated on the horizontal region of $I_1$, aligning better with the specified region in $I_0$. This indicates that the S6 model can better capture dynamics even with significant motion between frames.

\subsection{More qualitative comparison} 
\begin{figure*}[t]
    \centering
    \includegraphics[width=\linewidth, trim=1.5cm 12.7cm 2cm 3cm, clip=true]{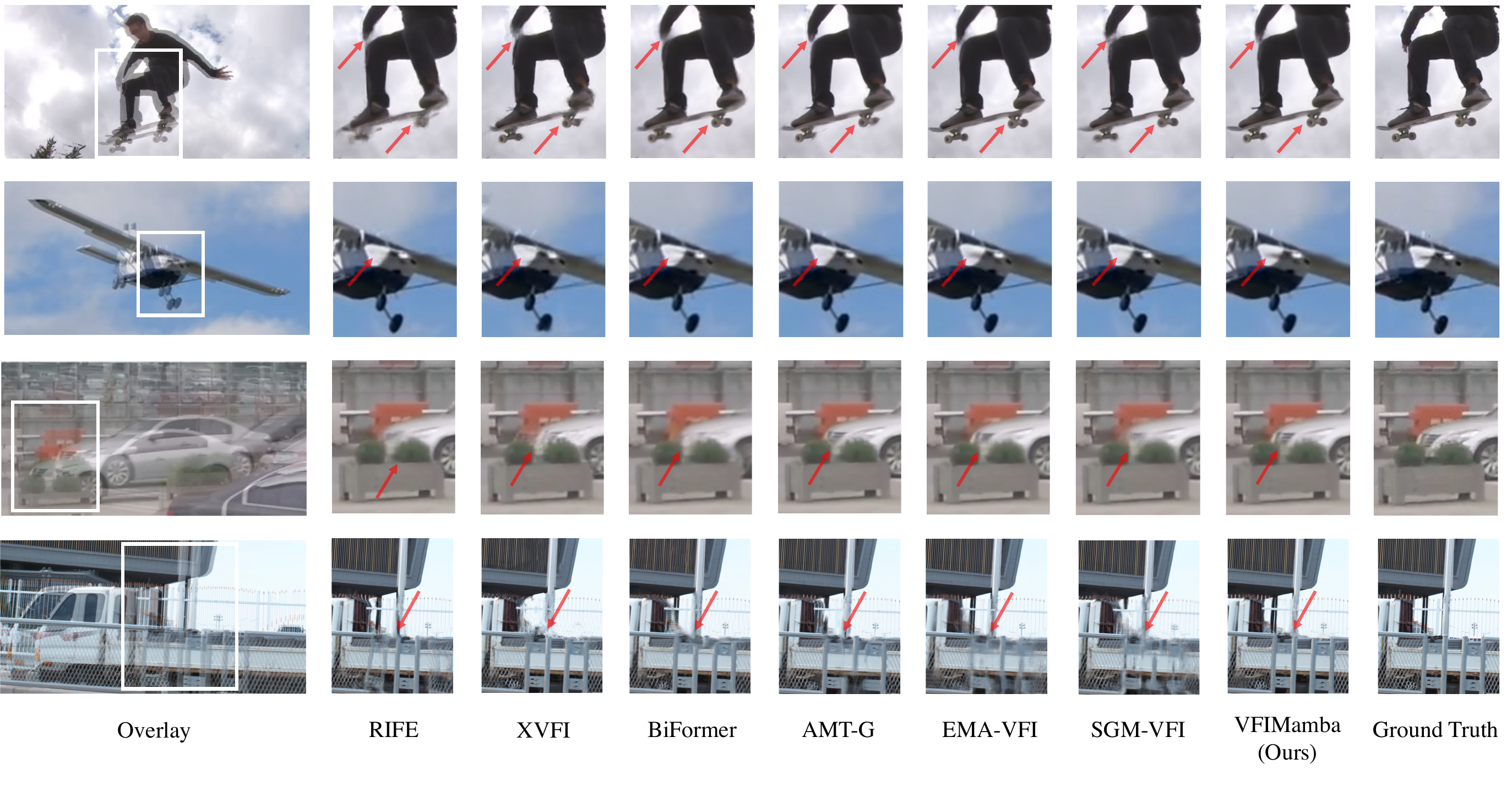}
    \caption{More Visualizations from SNU-FILM~\citep{reda2022film} and X-TEST~\citep{sim2021xvfi}.} 
    \label{fig:vis2}
\end{figure*}

As shown in Figure~\ref{fig:vis2}, we provide more visualization comparisons. VFIMamba demonstrates better visual quality compared to other methods.

\subsection{Model details}
\label{sm:model}

\subsubsection{Frame feature extraction}
\begin{figure*}[t]
    \centering
    \includegraphics[width=0.7\linewidth, trim=13cm 2cm 13cm 4cm, clip=true]{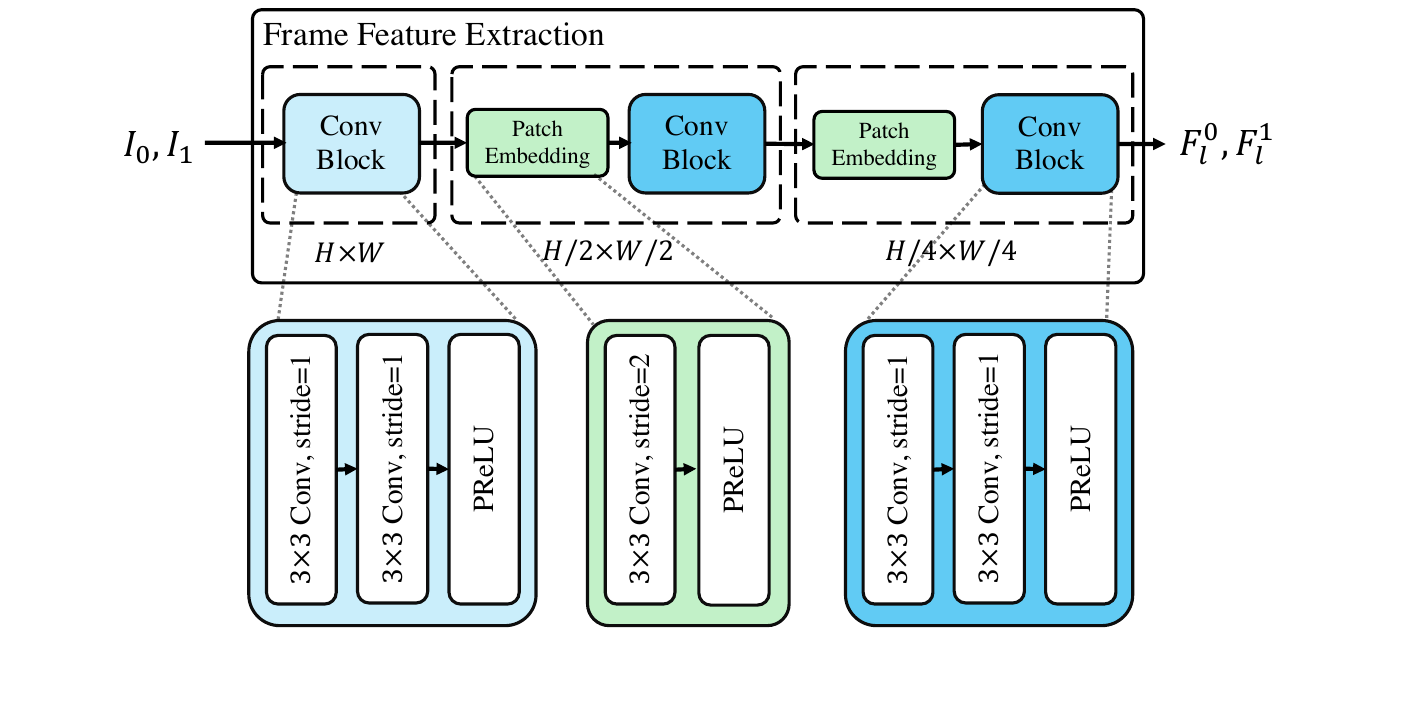}
    \caption{
    Details of frame feature extraction. The same color represents the same block structure.}
    \label{fig:featextract}
    \vspace{-2mm}
\end{figure*}
As shown in Figure~\ref{fig:featextract}, our frame feature extraction consists of multiple convolutional layers and PReLU~\citep{he2015delving}. The first convolution maps the image from 3 channels to $C$, with $C=16$ for VFIMamba-S and $C = 32$ for VFIMamba. Each time patch embedding is applied, the image resolution is halved, and the number of channels is doubled. Finally, we obtain the shallow features $F_l^i$ for each frame.

\subsubsection{Frame generation}
\begin{figure*}[t]
    \centering
    \includegraphics[width=\linewidth, trim=5cm 0cm 4cm 3cm, clip=true]{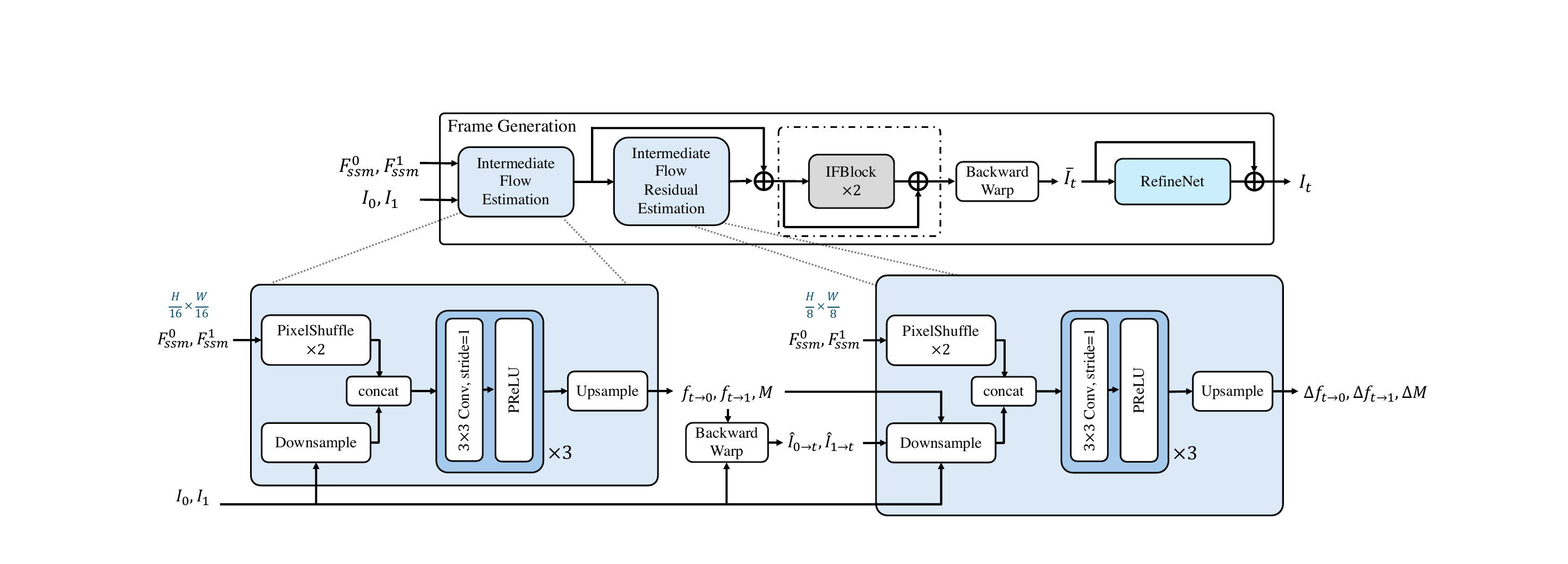}
    \caption{
    Details of frame generation. IFBlock is adopted from \citet{huang2022rife}, which is used optionally to enhance the local detail generation performance.}
    \label{fig:framegen}
    \vspace{-2mm}
\end{figure*}

As depicted in Figure~\ref{fig:framegen}, our frame generation includes an iterative intermediate flow estimation, local flow refinement, and appearance refinement using RefinNet. First, the intermediate flow estimation module uses the features $F_{ssm}^i$ obtained from inter-frame modeling with MSB for rough flow estimation. Specifically, we follow the design of EMA-VFI~\citep{zhang2023ema}, first utilizing features $F_{ssm}^i$ from the $\frac{H}{16}\times \frac{W}{16}$ scale and the original image $I_i$ for predicting the flow $f$ and occlusion mask $M$ by several convolutional layers. Then, we iteratively estimate the flow residual $\Delta f$ and mask residual $\Delta M$ using the $F_{ssm}^i$ from the $\frac{H}{8}\times \frac{W}{8}$ scale. After that, inspired by \citet{jia2022ncm}, which recognizes that the flow obtained through global inter-frame modeling may be coarse for high-resolution or large-motion scenes, we also introduce the IFBlock~\citep{huang2022rife} to further enhance flow accuracy in local details. We then use the predicted motion to backward warp~\citep{huang2022rife} the input frames to get the coarse intermediate frame $\bar{I}_t$. Finally, we adopt a U-Net-like~\citep{ronneberger2015unet} structure to predict the appearance residual using shallow features $F_l^i$ and inter-frame features $F_{ssm}^i$, resulting in the final frame $I_t$.

\subsection{Training details}
\label{sm:details}

\paragraph{Training loss} We used the same training loss as \citet{zhang2023ema}, which is a weighted combination of Laplacian loss~\citep{niklaus2020softmax} and warp loss~\citep{liu2019selflow}, with weights of 1 and 0.5, respectively.

\paragraph{Training setting} We used curriculum learning to train our model. For the data from Vimeo90K~\citep{xue2019vimeo}, we randomly cropped the frames from $256 \times 448$ to $256 \times 256$. For the data from X-TRAIN~\citep{sim2021xvfi}, since each sample contains 64 consecutive frames, we first randomly select two frames, starting with an interval of 1, which doubles every 50 epochs. Then, we randomly resized the frames from $512 \times 512$ to $S \times S$, where $S$ is initially 256 and increased by a factor of 1.1 every 50 epochs, and finally cropped them to 256 $\times 256$ for alignment. The larger the resize size, the greater the motion magnitude of the generated data. The batch size for Vimeo90K is 32, and for X-TRAIN it is 8. We then applied time reversal and random rotation augmentations. We used AdamW as our optimizer with $\beta_1 = 0.9$, $\beta_2 = 0.999$, and a weight decay of $1 \times 10^{-4}$. With warmup for 2,000 steps, the learning rate was gradually increased to $2 \times 10^{-4}$, and then we used cosine annealing for 300 epochs to reduce the learning rate from $2 \times 10^{-4}$ to $2 \times 10^{-5}$. Following \citet{jia2022ncm,park2023biformer,liu2024sgm}, we also trained the IFBlock separately on Vimeo90K for 100 epochs with same training setting to further improve the accuracy of local optical flow at high resolutions. The same procedure is followed for all ablation experiments.

\paragraph{Training time}
VFIMamba and VFIMamba-S were both trained on 4 NVIDIA 32GB V100 GPUs. Training VFIMamba-S takes about 38 hours, while training VFIMamba takes about 108 hours.

\subsection{Evaluation protocols}
\label{sm:eval}
In our paper, we primarily evaluated our methods on six benchmarks in terms of PSNR/SSIM\citep{wang2004image}: Vimeo90K~\citep{xue2019vimeo}, UCF101~\cite{soomro2012ucf101}, SNU-FILM~\citep{choi2020channel}, Xiph~\citep{xphi1994}, X-TEST~\citep{sim2021xvfi}, and X-TEST-L~\citep{liu2024sgm}. 

We followed the test procedures of \citet{huang2022rife} for Vimeo90K and UCF101, \citet{kong2022ifrnet} for SNU-FILM, \citet{niklaus2020softmax} for Xiph, \citet{reda2022film} for X-TEST with iterative $8\times$ frame interpolation, and \citet{liu2024sgm} for X-TEST-L with largest interval interpolation.

\subsection{License of datasets and pre-trained models}
\label{sm:license}
All the dataset we used in the paper are commonly used datasets for academic purpose. All the licenses of the used benchmark, codes, and pretrained models are listed in Table~\ref{tab:license}.

\input{Sections/X_license}

%% file: Sections/X_license.tex
\begin{table}
\centering
\caption{Licenses and URLs for every benchmark, code, and pretrained models used in this paper.}
\label{tab:license}
\resizebox{\linewidth}{!}{%
\begin{tabular}{cccc} 
\toprule
\multicolumn{2}{c}{Assets}                                                                         & License                                          & URL                                                                                                          \\ 
\midrule
\multirow{5}{*}{Benchmarks}                                                           & Vimeo90K   & MIT license                                      & \textcolor{blue}{\textcolor[rgb]{0.09,0.361,0.922}{https://github.com/anchen1011/toflow}}                    \\
                                                                                      & UCF101     & non-commercial research and educational purposes & \textcolor{blue}{\textcolor[rgb]{0.09,0.361,0.922}{https://github.com/lxx1991/pytorch-voxel-flow}}           \\
                                                                                      & SNU-FILM   & MIT license                                      & \textcolor{blue}{\textcolor[rgb]{0.09,0.361,0.922}{https://github.com/myungsub/CAIN}}                        \\
                                                                                      & XTEST (-L) & for research and education only                  & \textcolor{blue}{\textcolor[rgb]{0.09,0.361,0.922}{https://github.com/JihyongOh/XVFI}}                       \\
                                                                                      & Xiph       & freely redistributable                           & \textcolor{blue}{\textcolor[rgb]{0.09,0.361,0.922}{https://media.xiph.org/video/derf/}}                      \\ 
\midrule
\multirow{8}{*}{\begin{tabular}[c]{@{}c@{}}Codes and\\Pretrained Models\end{tabular}} & XVFI       & for research and education only                  & \textcolor{blue}{\textcolor[rgb]{0.09,0.361,0.922}{https://github.com/JihyongOh/XVFI}}                       \\
                                                                                      & FILM       & Apache-2.0 license                               & \textcolor{blue}{\textcolor[rgb]{0.09,0.361,0.922}{https://github.com/google-research/frame-interpolation}}  \\
                                                                                      & RIFE       & MIT license                                      & \textcolor{blue}{\textcolor[rgb]{0.09,0.361,0.922}{https://github.com/hzwer/ECCV2022-RIFE}}                  \\
                                                                                      & IFRNet     & MIT license                                      & \textcolor{blue}{\textcolor[rgb]{0.09,0.361,0.922}{https://github.com/ltkong218/IFRNet}}                     \\
                                                                                      & BiFormer   & Apache-2.0 license                               & \textcolor{blue}{\textcolor[rgb]{0.09,0.361,0.922}{https://github.com/JunHeum/BiFormer}}                     \\
                                                                                      & EMA-VFI    & Apache-2.0 license                               & \textcolor{blue}{\textcolor[rgb]{0.09,0.361,0.922}{https://github.com/MCG-NJU/EMA-VFI}}                      \\
                                                                                      & AMT        & CC BY-NC 4.0                                     & \textcolor{blue}{\textcolor[rgb]{0.09,0.361,0.922}{https://github.com/MCG-NKU/AMT?tab=License-1-ov-file}}    \\
                                                                                      & SGM-VFI    & Apache-2.0 license                               & \textcolor{blue}{\textcolor[rgb]{0.09,0.361,0.922}{https://github.com/MCG-NJU/SGM-VFI}}                      \\
\bottomrule
\end{tabular}
}
\end{table}